\documentclass[conference]{IEEEtran}
\usepackage{graphicx}
\usepackage{listings}
\usepackage{fancyvrb}
\usepackage{listings}
\usepackage{color}
\usepackage{natbib}
\usepackage[section]{placeins}
\graphicspath{ {images/} }
\definecolor{keywords}{RGB}{255,0,90}
\definecolor{comments}{RGB}{0,0,113}
\definecolor{red}{RGB}{160,0,0}
\definecolor{green}{RGB}{0,150,0}
 
\makeatletter
\def\@normalsize{\@setsize\normalsize{12pt}\xpt\@xpt
\abovedisplayskip 10pt plus2pt minus5pt\belowdisplayskip \abovedisplayskip
\abovedisplayshortskip \z@ plus3pt\belowdisplayshortskip 6pt plus3pt
minus3pt\let\@listi\@listI} 

\def\subsize{\@setsize\subsize{12pt}\xipt\@xipt}

\def\section{\@startsection {section}{1}{\z@}{24pt plus 2pt minus 2pt}
{12pt plus 2pt minus 2pt}{\large\bf}}

\def\subsection{\@startsection {subsection}{2}{\z@}{12pt plus 2pt minus 2pt}
{12pt plus 2pt minus 2pt}{\subsize\bf}}
\makeatother

\begin{document}
\date{}
\title{Product Review Based on Optimized Facial Expression Detection }

\author{
    \IEEEauthorblockN{
        Vikrant Chaugule, 
        Abhishek D, 
        Aadheeshwar Vijayakumar, 
        Pravin Bhaskar Ramteke, 
        and Shashidhar G. Koolagudi
    }
    \IEEEauthorblockA{
        \textit{National Institute of Technology Karnataka}\\
        Surathkal, Karnataka, India \\
        \{vikrant.chaugule, abhid95, vkaadheeshwar95, ramteke0001\}@gmail.com, koolagudi@nitk.ac.in
    }
}
%\end{tabular}}

\maketitle

%I don't know why I have to reset thispagesyle, but otherwise get page numbers
\thispagestyle{empty}

% Add this line right here!
\let\thefootnote\relax\footnotetext{This paper was published in the 2016 Ninth International Conference on Contemporary Computing (IC3). \copyright 2016 IEEE. DOI: 10.1109/IC3.2016.7880213}

\subsection*{\centering Abstract}
%IEEE allows italicized abstract
{\em
This paper proposes a method to review public acceptance of products based on their brand by analyzing the facial expression of the customer intending to buy the product from a supermarket or hypermarket. In such cases, facial expression recognition plays a significant role in product review. Here, facial expression detection is performed by extracting feature points using a modified Harris algorithm. The modified Harris algorithm reduced the time complexity of the existing feature extraction Harris Algorithm. A comparison of time complexities of existing algorithms is done with proposed algorithm. The algorithm proved to be significantly faster and nearly accurate for the needed application by reducing the time complexity for corner points detection.
}
\\\\
\begin{IEEEkeywords} 
	Gaussian Filter, Feature Extraction, Emotion Detection, Product Review.
 \end{IEEEkeywords}
\section{INTRODUCTION}
	Human facial expressions are having the strength to express feelings and emotions and control the behaviour among themselves. And facial emotion is an adaptive response which would give a picture of mental state of the person (consumer in particular). The indication of many emotions like happy, sad, interested, curious, excited etc can be done via facial expressions. The fast growing technology has led to the communication between humans and robots, and robots can now be instructed through facial expressions. Machines recognising facial expression of human has been an active area of research. The major problem in facial expression recognition is handling variations in expressions, processing faster and also product specific development. Some products need faster algorithms and some accurate algorithms, and balancing between these two factors is the major concern\cite{de1997facial}. Emotion detection has many applications in practice, and one of them is product review. For example, the statistics on the consumers curiousness can be a hint for "how much interested they are in the product".
 
	One very good example one can cite when one talk about product review is the Ponds toothpaste. Ponds is a product of the very well established and renowned company Hindustan Lever which was founded in 1885. The flagship brands of the company like ‘Ponds’ and ‘Clinic All Clear’ were very successful and had expanded their operations in manufacturing soaps, face wash, age defying creams etc.. However, they were not very careful in which territory they tread and decided to come up with a toothpaste. ‘Ponds’ which was a popular face cream brand had failed miserably when it applied its name to toothpaste. Since for customers, the brand Ponds stood for nothing but fragrance and a product for external use, it missed the required attributes of  a toothpaste which is taste. This lead to an instant failure of the product since consumers was very sceptical about buying the product itself. It also led to a lot of loss and damage to their brand image since they realized this a bit late. Hence, in this case, the first reaction of the consumer on seeing the product would have been very valuable and could have saved the company from facing this damage.

The paper’s primary focuses are as follows. Sub Division 2 illustrates the background knowledge of existing algorithms for facial detection, and feature points extraction. Sub Division 3 illustrates the problem statement. Sub Division 4 presents the proposed model and technique of product review Sub Division 5 explains in detailed implementation of the approach along with case study. Sub Division 6 provides with the Evaluation of proposed approach through survey. Sub Division 7 concludes our work and also the future work needed
  
\section{LITERATURE REVIEW}
The Literature survey has been done on the methods and algorithms which are already in use for face detection, emotion detection and feature points detection and extraction.
\subsection{FACE DETECTION}
	The proposed algorithm applies the method of face detection on the basis of Viola John algorithm. The inner details of this algorithm is briefed using Integral Image, Adaboost algorithm and Haar like feature to convert weak classifier into a stronger one and to get higher performance than existing face detection algorithms.   
	
\subsubsection{Integral Image} 

Viola John used this for the first time for image processing and it is the intermediate representation of given image. Using this the calculation of integral image rectangle feature became very fast\cite{de1997facial}.

 The integral image at location  p, q has the sum of the pixels above and to the left of  p, q inclusive:  
   \begin{equation}
   	ii(p,q) = \sum_{p1<=p, q1<=q} i(p1, q1);
   \end{equation}
where y(p, q)  is the integral image and i(p, q) is original image. The following pair of recurrences are used to compute the Integral Image in one pass. 
\begin{equation}
	x(p, q)  =  x(p, q-1) + i(p, q)
\end{equation}
\begin{equation}
	ii(p, q) =  ii(p-1, q) + x(p, q)
\end{equation}

where x(p, q)=cumulative row sum, x(p, -1) = 0 and y(-1, q) = 0. 

\subsubsection{Haar Like Feature}

Before this feature was introduced the feature calculation was costly to compute. These features are like wavelet Haar features. One can get these features in the form of intensity through integral image method upon given original image. Pixel sum of white region subtracts pixel sum of black region. But the features calculated (in subwindows) are very larger than pixels itself. To avoid this and to reduce the complexity of feature selection the Adaboost algorithm is used. It is shown in figure 1. 

\begin{figure}[htb]
\caption{Haar like feature}
\includegraphics[scale=0.7]{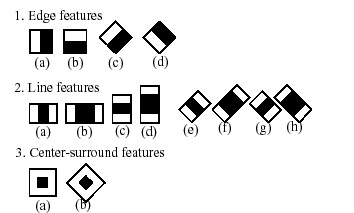}
\end{figure}

\subsubsection{Adaboost} 

Adaboost for Haar cascade method is used to train by supervised learning to classify negative and positive sample, to classify the results and show how AdaBoost learns weak classifier by converting weak classifier to a stronger classifier and reducing the computation time. At all the stages of cascade, application of more strict rules for adding more or less different Haar feature and the other features are rejected. 

When an image is passed to a classifier cascade and if the image passes all classifiers then there will be a very high probability that face is present and is shown in figure 2.           

\begin{figure*}[htb]
\caption{Cascade of feature Classifiers}
\includegraphics[scale=0.8]{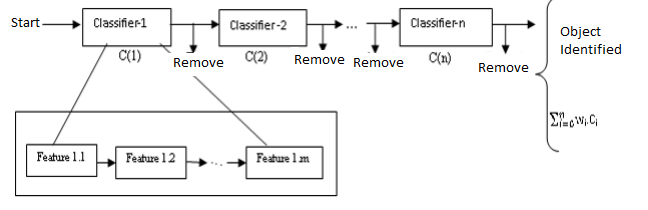}
\end{figure*}

Mouth detection-Mouth is detected in same manner as the face is detected.
Eye detection-Eye is detected in same manner as the face is detected.

\subsection{EMOTION DETECTION}
Emotion recognition is the process of identifying human emotion, most typically from facial expressions. This is both something that humans do automatically but computational methodologies have also been developed.
Some of the oftenly used terminologies in the paper related to emotion detection are briefed here.
\subsubsection{Gaussian filter}
 
 The Operator used for Gaussian filtering does a weighted average of pixels surrounding it based upon Gaussian distribution. This is used in removing Gaussian noise and is a model of realistic defocused lens. 
Sigma defines amount of blurness. The radius slider controls the largeness of the template. The larger the sigma is larger will be the blurring for larger sizes.  
How it works is explained below. 
 
The Gaussian operator generates a value template that are applied then to the pixel group in the image. Theses values of templates are defined by the equation given below and shown in figure 3. 
 
 \begin{equation}
 	 w_{u,v}=exp(-\frac{1}{2}(u^{2}+v^{2})/\delta ^{2})
  \end{equation}
\begin{figure}[htb]
\caption{Graph showing high sigma values require significantly more calculations per pixel}
\includegraphics[scale=0.8]{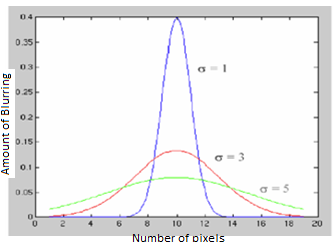}
\end{figure}

The figure shows variation in curve in accordance with sigma values.
 
 \subsubsection{Beizer curve}  
 
 In computer graphics the Bezier curves are majorly used for modelling smooth curves. The curve can be manipulated intuitively and the points can also be shown graphically as convex hull of control points contains the curve. Translation and rotation is applicable to the curve by doing the respective transformations over the control points of the curve.
 For rasterizing this curve the simple method would be evaluating it in most of the points closely spaced and converting the approximate line segment sequence\cite{mcclure2003facial}.

\subsection{FEATURE POINTS DETECTION AND EXTRACTION}
Numerous applications require relating two or more pictures to concentrate data from them. For instance, if two progressive edges in a video succession taken from a moving camera can be connected, it is conceivable to extricate data in regards to the profundity of items in the earth and the pace of the camera. The brute force technique for contrasting each pixel in the two pictures is computationally restrictive. Naturally, one can picture relating two pictures by coordinating just areas in the picture that are somehow intriguing. Such points are alluded as interest points and are found utilizing an interest point identifier. Finding a relationship between pictures is then performed utilizing just these points. This definitely lessens the required calculation time. Various interest point locators have been proposed with an extensive variety of definitions for what focuses in a picture are fascinating. A few indicators discover purposes of high neighbourhood symmetry, others discover territories of exceedingly fluctuating composition, while others find corner points.

Corner points are fascinating as they are shaped from two or more edges and edges more often than not characterize the limit between two unique questions or parts of the same item. There are different techniques to extricate the corner points (interest points), however each has its own particular advantages as well as drawbacks.
 \begin{figure*}[htb]
 \caption{flow chart of the proposed approach}
\includegraphics[scale=0.63]{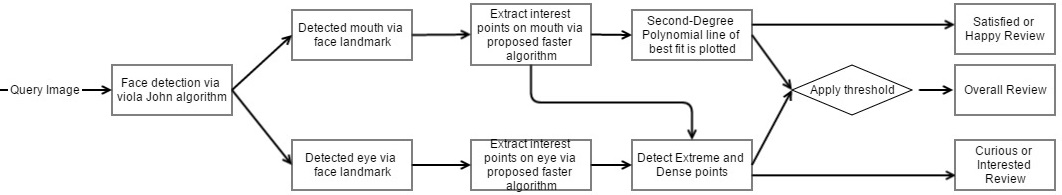}
\end{figure*}
\subsubsection{Susan corner detector} S.M. Smith and M. Brady presented SUSAN operator, and discover the interest point by contrasting nucleus pixel with all pixel inside the circular mask, center  pixel have intensity  more  than every one of the pixels in the mask like center  of gravity principles and in the event that we expand limit at certain point it identifies edges rather than corner\cite{zhao2008adaptive}. 

\subsubsection{Rostern \& Drummond corner detector}

It  distinguishes the corner through local intensity comparisons of the pixels and is the  quickest corner identifier which takes less time to the various techniques though keeping accuracy at stake(accuracy is reduced by half). 

\subsubsection{Harris \& Stephens corner detector}

A selection criteria is used for the detection of corner points. For every pixel, a score is resolved and if this score is greater than the certain value, the pixel is considered as a corner. This algorithm is clarified mathematically in the implementation segment recommending the modifications that the proposed algorithm does\cite{harris1988combined}.

\subsubsection{Shi \& Tomasi}

This algorithm  is based on the Harris Stephens corner detection. The primary difference between the two approaches are that, with Shi and Tomasi, the corner is detected by calculating minimum of two eigenvalues of the matrix rather than computing  the score from the function \cite{kovesi2000matlab}.
 
The paper concentrates on a problem which can be divided into two parts. The first problem is that companies take a lot of time to get a feedback of their product. Usually by the time they get a public opinion of their product the brand image of the company goes down. Also for new products there is a delay involved in finding out the consumers perception of the product launched. The proposed approach helps companies to take up quick action to rectify the shortcomings of the product. The proposed approach helps in reviewing product (especially for newly launched products in the market) by analysing the facial emotion of all the costumers viewing the product. Hence we intend to give data chart for the company which should provide them the statistics regarding general trend in consumer's mind pattern. The second problem is that the existing emotion detection algorithms use predefined feature extractors which are slow which might affect the time required to perform Emotion analysis using real time video. The challenge intended to be solved is to accelerate the process of emotion detection such that it is suitable for real time video and also has nearly the same accuracy as previous algorithms\cite{kovesi2000matlab}.

\section{METHODOLOGY}

	The proposed approach takes a  real time video or recorded video as input and takes snapshots at regular intervals. In advanced commercial deployments, the insights extracted from these frames can be fed into generative AI frameworks for dynamic video trailer synthesis \cite{dharmaratnakar2026generative}, allowing product advertisements to automatically adapt to viewer reactions. Furthermore, this facial analysis pipeline could be embedded within highly interactive consumer environments, such as those built using Domain Specific Language (DSL) architectures for gaming applications \cite{vijayakumar2016dsl}. These snapshots are then analyzed. The face is then detected using Viola-John Algorithm, which is also separated from the snapshot. The face is then  processed to detect mouth accurately by only considering lower one-third of the detected face. Similarly, eyes are detected by considering the upper one-third of the face. Then we apply the proposed approach of finding feature points which is done in a much  faster manner such  that nearly five times more snapshots can be processed and analyzed compared to regular methods used previously. The feature points extracted from both mouth and eyes are then processed (detailed explanation in implementation).The  review of the customer is calculated using the curious ratio calculated using eyes and mouth as well as the curve obtained using the feature points of the mouth. The algorithm gives appropriate reviews by giving proper weights to each of the values extracted from eyes and mouth. From these values calculated, expressions like curiousness, dis-interestedness, satisfaction and the excitement level of a person can be identified\cite{shi1994good}. Hence, the overall rating is calculated for the product being viewed. The approach proposed can be represented in the form of a flow diagram. For clearer understanding, the input to the algorithm and output from the algorithm is shown as well\cite{viola2004robust}. 
The second part of the paper emphasizes on accelerating the process of emotion detection by going through the intricacies of existing algorithms( discussed in implementation ). The algorithm manages to keep nearly the  same accuracy as the existing algorithms while  reducing the time complexity using the approximation algorithmic techniques and is shown in figure 4.

\section{IMPLEMENTATION}
The implementation of the algorithm is based on the study of previous existing algorithms and is mostly based on algorithm by Harris. The Harris algorithm for feature extraction involves calculation of gaussian filter matrix which is the major time consuming part of the entire algorithm\cite{mao2009improved}. The functions such as convolution of matrices and multiplication of matrices can not be optimized further. The only change that the algorithm proposes is to reduce the time complexity by using taylor expansion of exponentiation and cutting down the existing complexity of the algorithm by \begin{math}log(n)\end{math}
\subsection{Product Review implementation}
The complete procedure is as described. Face is detected and
extracted via Viola Jones algorithm, which is considered
a standard algorithm for face detection. Then the
lower part of the detected face separated from which the mouth is extracted using  vision library of  MATLAB. Only the lower one-third of the face is analysed for better accuracy of mouth detection. Similarly upper one-third  is separated and eyes are extracted. The optimized algorithm is then  used to extract feature points(discussed later).  

 	In the next step the construction of Bezier curve from mouth points is done and then the derivative of the curve of the best fit is calculated and the points' concavity is found, accordingly results are obtained. 
	 Bezier Curve-The construction of Bezier curve can be done using four points. They are one starting point, two points for controlling and one ending point\cite{rhodes2006evolutionary}. But for this approach more than 4 points are taken and hence curve gives better results. Once the curve is found the MATLAB tool for curve fitting is used to fit the curve to a second degree of polynomial and a threshold is set which if met then only considered for various types of emotion detection.
 \begin{figure}[htb]
\caption{Corner Detection Theory}
\includegraphics[scale=0.5]{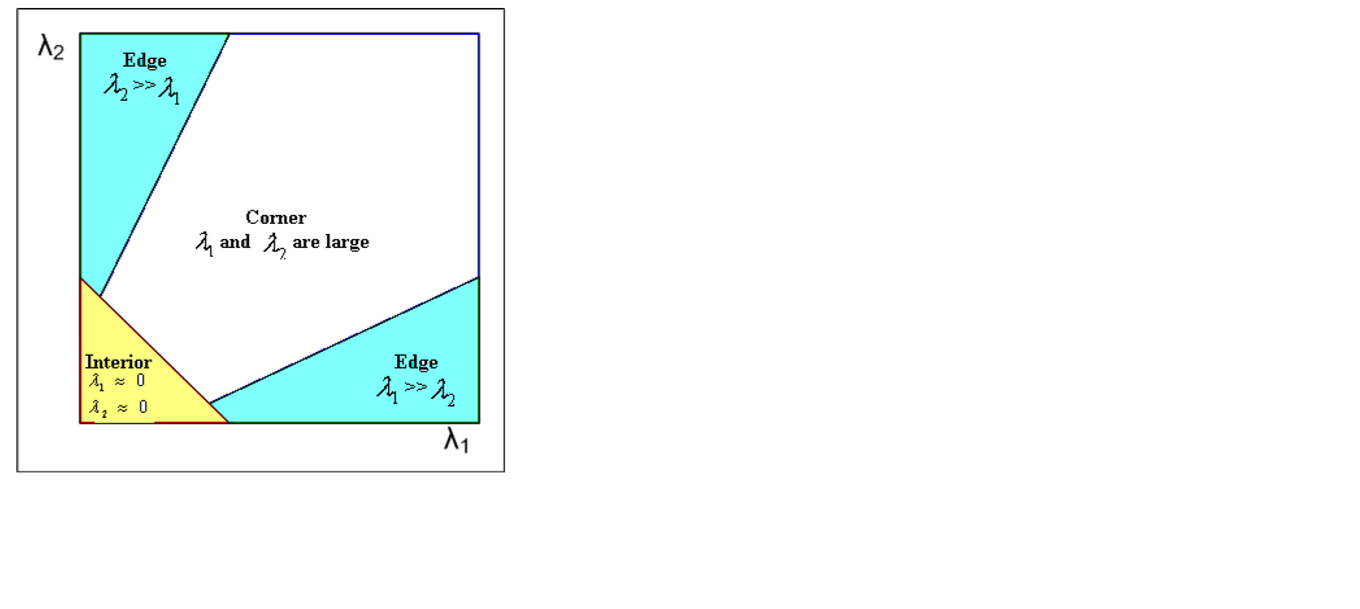}
\end{figure}					 

	. For the eyes, we iterate through all feature points y(j) for each x(i) and then maximum and minimum values are appropriately calculated. Using these values, the curious ratio for eyes is calculated. The same procedure followed for feature points of the mouth  and the curious ratio for mouth is calculated. Both of these values help in deciding the curiousness level of the person. Also if values of both mouth and eye curious ratio are above a threshold value (after application of proper weights), it helps decide the  excitement level of a person. In the same way the weights are assigned to all three factors i.e mouth curious ratio, eye curious ratio and smile curvature.If the overall value is less than a threshold value then it is assumed that  the person is disinterested in the particular
product. The overall product  rating is also calculated based on
the above mentioned factors\cite{durand2007development}.

 \subsection{Optimized feature extractor}
  %copy begin
  The algorithm used as a base is Harris algorithm which is explained in detail along with the change we are proposing.
  The signal point feature extraction is the basis of the algorithm. This algorithm makes the window of the image to move to infinity in any direction\cite{durand2007development}. Also the gray variation is defined as follows. 

 \begin{equation}
 E_{x,y} = \sum_{u,v}w_{u,v}[I_{x+u,y+v} - I{u,v}]^{2}
 \end{equation}
  \begin{equation}
 = \sum_{u,v}w_{u,v}[xX + yY + O(x^{2} + y^{2})]^{2}
 \end{equation}
  \begin{equation}
 = Ax^{2}+2Cxy+By^{2}
 \end{equation}
  \begin{equation}
 = (x , y)M(x , y)^{T}
 \end{equation}
	where X, Y are the 1-order convoluted gray-level gradient of the image. 
  \begin{equation}
	X = \frac{\partial I}{\partial x} 
		= I\otimes (-1,0,1)
  \end{equation}
  \begin{equation}
	Y = \frac{\partial I}{\partial y} 
	= I\otimes (-1,0,1)
  \end{equation}
  To improve the capabilities of anti-noising, Gaussian window is used to smooth image window\cite{vedaldi2010vlfeat}.
  \begin{equation}
 	 w_{u,v}=exp(-\frac{1}{2}(u^{2}+v^{2})/\delta ^{2})
  \end{equation}
	 %copy end
  This is where the gaussian function can be optimized and taylor expansion upto 3 to 4 terms can be applied ( as remaining terms only contribues to 2\% of the entire result ) for exponentiation function used to find gauss value. So the algorithm detects feature points by reducing the complexity of calculation gauss value atleast by log(MAX) times. (MAX is the maximum value in x and y matrices)
  \begin{equation}
  e^{x} = 1+\frac{x^{1}}{1!}+\frac{x^{2}}{2!}+\frac{x^{3}}{3!}+\frac{x^{4}}{4!}+\frac{x^{5}}{5!}+...
  \end{equation}
   %copy begin
  And define: 
  \begin{equation}
  A=X^{2}\otimes w, B=Y^{2}\otimes w, C=(XY)\otimes w.
	\end{equation}
	
It encompasses the function for partial autocorrelation, and at the origin the autocorrelation function shape is described by M.    

  Encloses to the partial autocorrelation function, and M describes the shape of the autocorrelation function at the origin\cite{huang2009face}. M's Eigen values will be $\lambda_{1}$ , $\lambda_{2}$ . 
And $\lambda_{1}$
, $\lambda_{2}$
are in proportion to the 
are proportional to the curvature of partial autocorrelation function's curvature and rotational invariance for M is constituted by this\cite{trajkovic1998fast}.
 Then this can find the edge, corner point, the flat area, by $\lambda_{1}$ and $\lambda_{2}$ values ,The following three conditions are possible.
\\
a. The flat autocorrelation function is shown if 2 curvatures' values are small
\\
b. The ridge-like autocorrelation function and small change in E is shown if one curvature is small and other is large. 
\\
c. A peak is found in autocorrelation function and a large change in E in any direction if both the curvatures are large.
\\
This analysis and division can be described by this figure below\cite{wang1995real}.

 %write above statements properly
 Then, the feature point is the maximum value of local area: 
 %eqns
 \begin{equation}
 	Det(M) - k Tr^{2}(M)
 \end{equation}
 \begin{equation}
 	Tr(M) = \lambda_{1} + \lambda_{2} = A + B
 \end{equation}	
 \begin{equation}
 	Det(M) = \lambda_{1}\lambda_{2} = AB - C^{2}
 \end{equation}
 
 Where $Tr(M)$ is the trace of 
matrix M ; $Det(M)$ is the determinant value of matrix M ; the best results are given by k=0.04 to 0.06 values\cite{dixit2013review}.
  %copy end
  \begin{figure}[htb]
\caption{Comparison on happy face}
\includegraphics[scale=0.6]{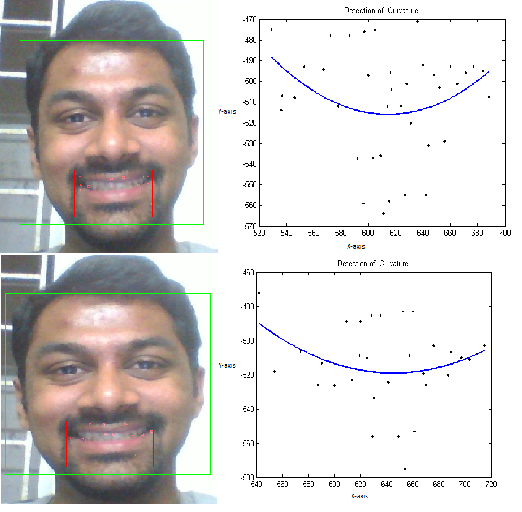}
\end{figure}

\begin{figure}[htb]
\caption{Comparison on sad face}
\includegraphics[scale=0.4]{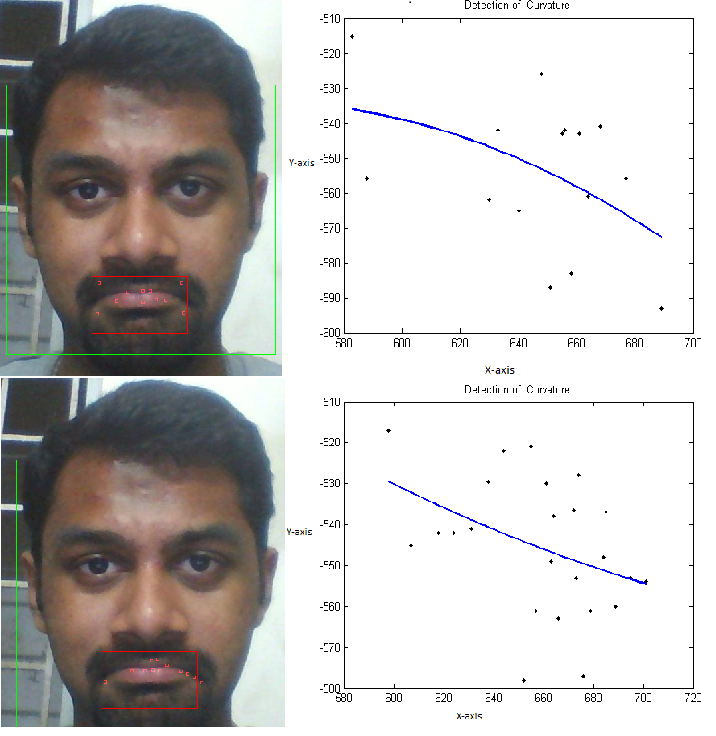}
\end{figure}

\section{Evaluation of proposed approach (Survey)}

 The approach for product review using emotion detection has been done properly for the first time and hence comparative study is not required. The comparison for feature extraction algorithm over the previous one is required as its a new approach suitable for product review and known for its speed. These following results are verified by running over genki dataset.

\subsection{Time required for processing}

 The Harris algorithm was using $n^{2}log(MAX)$ whereas in the newly proposed algorithm we propose to do it in $n^{2}$ time by reducing the time used to calculate gaussian function. Below is the table of time taken by older and proposed approaches run over a video taking 100 snapshots at regular intervals of time. 
 And in the graph we can see that as Max value and size of matrix increases the time taken also increases drastically for older algorithms and in newer algorithm its almost taking 2 to 3 times lesser time.
 \begin{figure*}[htb]
\caption{Table showing time required for processing}
\includegraphics[scale=0.8]{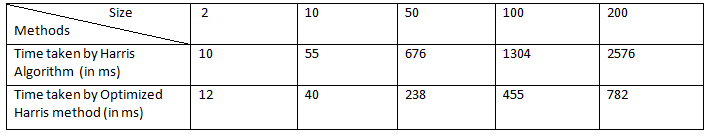}
\end{figure*}  
\begin{figure}[htb]
\caption{Graph showing time required for processing}
\includegraphics[scale=0.5]{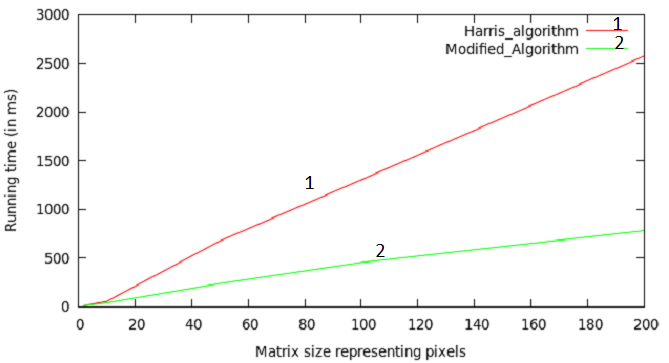}
\end{figure}
 The Harris algorithm uses proper exponential function whose worst case complexity is log(MAX) but by reducing it using approximation algorithm we are able to extract features with almost same accuracy. Comparison table has been provided for same parameters in previous table and results are plotted. 
\iffalse
In this survey comparison is made between the manual coding and automated code generation by DSL for various sites involving HTML source codes with different number of features(classes). It can be observed that automated code generation saves a lot of time than manual coding\cite{rizvi2002maintaining}. Table 2 gives the picture of it.
Here manual code writing part is not the one taking the time. Rather it is time consuming to study the whole source code and finding out the tag and class fields, which is automated in the proposed approach.

\begin{figure}[htb]
\caption{Graph showing time to study webpage and code in older and newer ways}
\includegraphics[scale=0.5]{time_graph.png}
\end{figure}  
\fi

\subsection{Comparison of accuracy over previous algorithms}
The Harris algorithm uses proper exponential function whose worst case complexity is log(MAX) but by reducing it using approximation algorithm we are able to extract features with almost same accuracy. Comparison table has been provided for same parameters in previous table and results are plotted which is done in figure 6 and 7. 
 \begin{figure*}[htb]
\caption{Table showing comparison of accuracy over previous algorithm}
\includegraphics[scale=0.8]{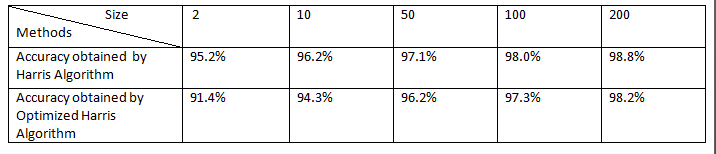}
\end{figure*}  
\begin{figure}[htb]
\caption{Graph showing comparison of accuracy over previous algorithm}
\includegraphics[scale=0.5]{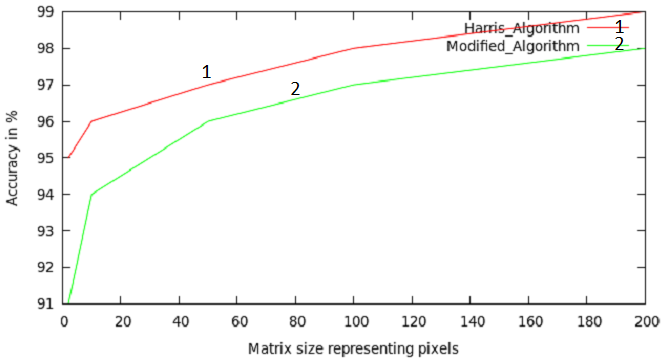}
\end{figure}
\iffalse
\section{CASE STUDY ON PRODUCT REVIEW}
	The case study we have taken is a study which was done in the sales of sunsilk BLACK in UK and other European nations. The story goes this way. The marketing of Sunsilk BLACK was first done in India and it was a big hit for the company as the people in India had black hair and considered black hair is the sign of youth or young age. As company made a huge profit in India they moved to European nations to market the product in huge amounts. But Europeans rarely has black hair and upon seeing the product black in colour they felt it would run down their face and turn their face black. Black hair is not considered as the sign of youth rather black is considered ugly. Hence the sunsilk corporation faced a huge loss because they got to know what people felt about the product after 2 to 3 months. 
	Instead if they had done product review based on the expression of people upon seeing the product. They could have got to know and would have stopped their production or modified their product. Such losses have been incurred by many other companies due to time lag in marketing and reviewing the product. Hence the proposed approach helps companies to keep track of the public opinion of the product from the first day of their product release into market.
	Also the fast feature extractor helps the proposed approach to process the real time video faster and thereby collecting more data by taking more snapshots.  
\fi
\section{CONCLUSION AND FUTURE WORK}
The proposed fast feature extractor helps process data faster thereby collecting more data and reducing queuing up of data(snapshots). Also the video after processing need not be stored as it can be done in real time and thereby protecting privacy and security of any consumer. Also large amount of video data storage for each and every product in a supermarket or online shopping sites is huge risk and investment on storage can be reduced to a large extent. So reduction in computational power, storage capacity, security and privacy of consumers are all taken care by the proposed approach. The future work may be to see through other factors along with emotions for product review like the time consumer spends to see a product and it requires us to do facial recognition as well to achieve this task in supermarkets, but it can be achieved easily as far as online shopping is concerned. 
Today everyone is drawn towards online shopping, such reviews may help in early prediction of failure of product preventing further loss in manufacturing. To elevate the e-commerce ecosystem, real-time facial expression data could act as a critical input signal for multi-agent video recommenders \cite{ranganathan2026multi}. By synergizing our optimized detection algorithm with reliable trajectories in agentic Information Retrieval (IR) systems \cite{sinha2026beyond}, platforms can dynamically serve highly personalized product recommendations based on instantaneous emotional feedback.
Finally, as consumer expression data becomes central to product analytics, ensuring the privacy of customer telemetry is paramount; future systems may necessitate secure data-logging techniques analogous to semantic text steganography \cite{vaishakh2018semantic}. Similarly, as these facial analysis metrics are increasingly processed by backend automated reasoning systems, mitigating false positives and ensuring algorithmic factuality much like the ongoing efforts to curb hallucinations in Large Language Models \cite{ranganathan2026factuality} will be essential for reliable market research.

\bibliographystyle{unsrt}
\bibliography{ref}
\end{document}